\documentclass[conference]{IEEEtran}
\IEEEoverridecommandlockouts
\usepackage{cite}
\usepackage{amsmath,amssymb,amsfonts}
\usepackage{algorithmic}
\usepackage{graphicx}
\usepackage{textcomp}
\usepackage{xcolor}
\usepackage{mdframed}
\newmdenv[
  backgroundcolor=gray!10,
  linecolor=gray!40,
  roundcorner=4pt,
  skipabove=6pt,
  skipbelow=6pt,
  innerleftmargin=8pt,
  innerrightmargin=8pt,
  innertopmargin=6pt,
  innerbottommargin=6pt
]{graybox}

\usepackage{lineno}
\usepackage{subcaption}
\usepackage{booktabs}
\usepackage{hyperref}
\usepackage{enumitem}

\def\BibTeX{{\rm B\kern-.05em{\sc i\kern-.025em b}\kern-.08em
    T\kern-.1667em\lower.7ex\hbox{E}\kern-.125emX}}
\begin{document}

\title{S2ED: From Story to Executable Descriptions for Consistency-Aware Story
Illustration}

\author{
\IEEEauthorblockN{
Sijing Yin\IEEEauthorrefmark{1}\IEEEauthorrefmark{2},
Jiamou Liu\IEEEauthorrefmark{1}\IEEEauthorrefmark{2}\IEEEauthorrefmark{3}\IEEEauthorrefmark{4},
Xiao Tang\IEEEauthorrefmark{1},
Yaser Shakib\IEEEauthorrefmark{2},
Qian Liu\IEEEauthorrefmark{1}
}
\IEEEauthorblockA{\IEEEauthorrefmark{1}University of Auckland}
\IEEEauthorblockA{\IEEEauthorrefmark{2}Bedaia.ai}
\IEEEauthorblockA{\IEEEauthorrefmark{3}Wuhan College of Communication}
\IEEEauthorblockA{\IEEEauthorrefmark{4}Corresponding author}
\IEEEauthorblockA{
syin565@aucklanduni.ac.nz, jiamou.liu@auckland.ac.nz
}
}

\maketitle
\begin{abstract}
Multi-frame story illustration requires long-horizon coherence beyond single-image text-to-image generation, including narrative decomposition and persistent character identity, layout, and affect across frames.
We propose \emph{Story-to-Executable Descriptions (S2ED)}, a training-free, model-agnostic, prompt-layer framework that converts a full story into a sequence of explicit, editable executable descriptions for more consistent rendering.
S2ED coordinates three agents to segment the narrative, ground canonical character attributes, and enrich spatial and affective cues, enabling interpretable prompt-carried state propagation and local edits to repair drift without retraining the generator.
Experiments on \textit{Flintstones} and \textit{Shakoo Maku} show that S2ED improves sequence-level consistency and character fidelity over strong prompting, large-model planning, and a reference training-based method, under both automatic metrics and human judgments.
We also deploy S2ED in an end-to-end story-to-storybook system for children's illustrated stories, with a supplementary video.

\end{abstract}

% \begin{IEEEkeywords}
% Story Visualization,
% Structured Prompting,
% Multi-Agent Systems,
% Character Consistency,
% Layout-Aware T2I,
% Diffusion Models.
% \end{IEEEkeywords}


\section{Introduction}
\label{sec:intro}

\begin{figure}[t]
  \centering
  \includegraphics[width=0.5\textwidth]{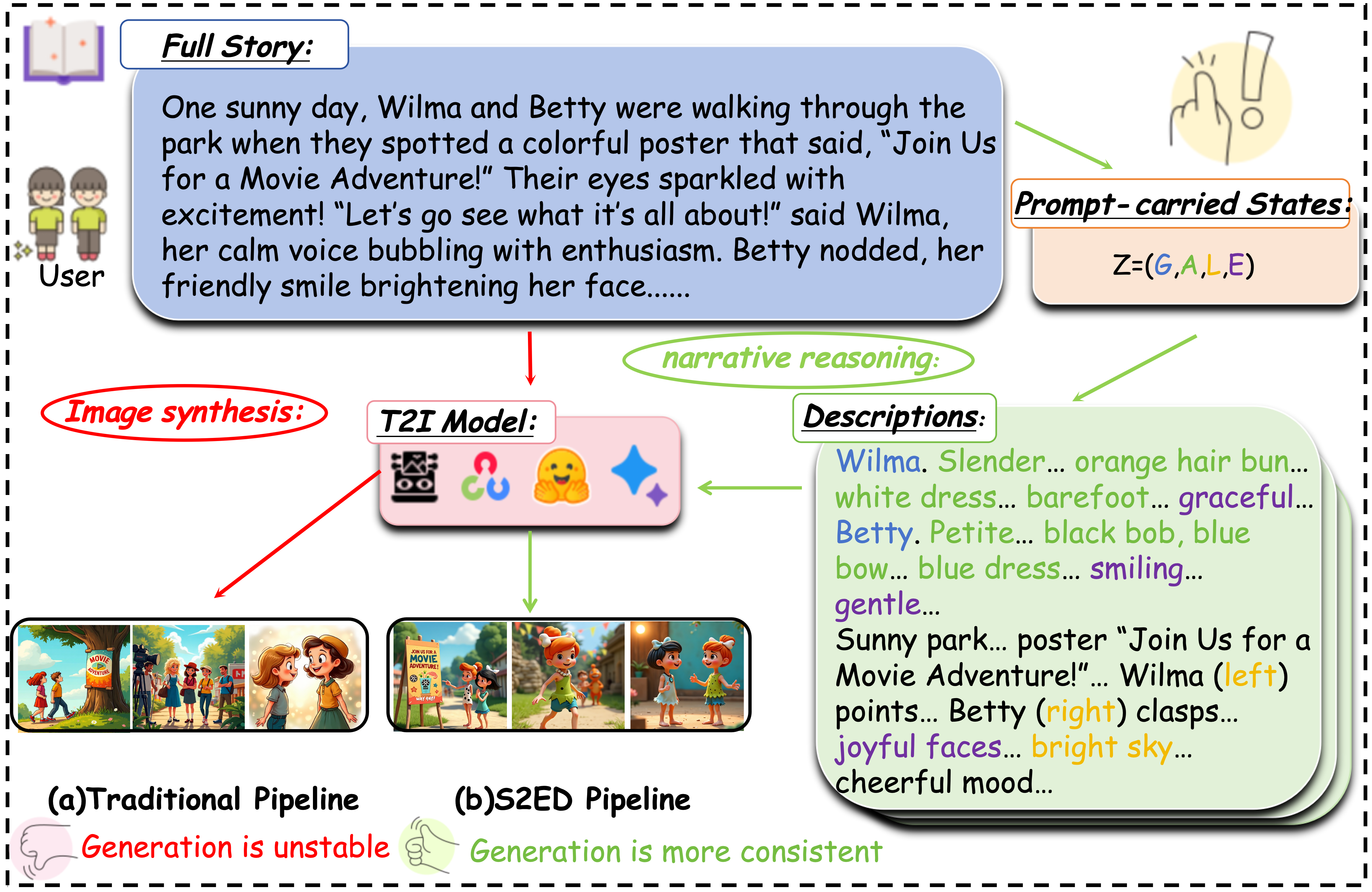}
  \caption{Unlike the traditional pipelines (red) which view story illustration primarily as a multi-frame image generation task, the S2ED pipeline (green) decouples narrative reasoning from image synthesis thru an explicit state interface.}
  \label{fig:introteaser}
\end{figure}

\emph{Multi-frame story illustration} converts a narrative into a coherent sequence of images and supports applications such as picture-book creation, assistive storytelling, and pre-visualization for media production~\cite{NEURIPS2024_c7138635,liang2024story}. 
However, unlike single-image generation, story illustration demands long-horizon planning and narrative pacing, fine-grained persistence of character identity, spatial layout, and affect across frames.
% As shown in Fig.~\ref{fig:introteaser}, standard prompting pipelines often generate panels in isolation, so attributes and composition are sampled conditioned only on its local prompt without sharing state across frames, producing drifts~\cite{NEURIPS2024_c7138635}.
As shown in Fig.~\ref{fig:introteaser}, traditional prompting pipelines generate each frame independently, conditioning image synthesis solely on frame-local text without an explicit, shared intermediate representation, which leads to uncontrolled drift in character attributes, layout, and affect across frames~\cite{NEURIPS2024_c7138635}.

%\emph{Multi-frame story illustration} aims to transform narrative text into a coherent sequence of images \cite{NEURIPS2024_c7138635}. 
%The task sits at the intersection of controllable text-to-image (T2I) generation and narrative understanding, and underpins a wide range of real-world applications, including picture-book publishing \cite{rexigel2024more}, accessibility and assistive storytelling tools \cite{das2024provenance}, pre-visualization for animation, film, and game production \cite{liang2024story}, as well as human--AI co-creative systems for visual storytelling \cite{moruzzi2024user,gmeiner2024evidence}. 
%Unlike single-image generation, story illustration demands long-horizon planning and narrative pacing, fine-grained persistence of character identity, spatial layout, and affect across panels, and interpretable representations that support human-in-the-loop control \cite{Storytelling2020}. 
%Without these capabilities, visual narratives often suffer from inconsistency and drift, undermining readability and user trust. 
% In this work, we go beyond conceptual relevance and demonstrate S2ED in a real-world children’s storybook generation system, where narrative consistency and editability are essential for practical use.

Existing remedies only partially address this issue. 
\emph{(1) Model retraining and fine-tuning}, such as DreamBooth \cite{ruiz2023dreambooth} or LoRA, stabilize identity but require per-subject data and high compute, and they reduce portability \cite{hu2021}. 
\emph{(2) Embedding and token injection} encode traits into special tokens. 
They improve single-frame identity but do not manage layout or affect across frames \cite{gal2022image}. 
\emph{(3) Prompt-template strategies} are most desirable in practical deployment as they are training-free, thereby avoiding  the cost, latency, and maintenance burden of retraining \cite{brooks2023instructpix2pix}. Such approaches include fixed schemas or one-prompt-for-all-frames. However, they break down in long stories due to token limits and the lack of state propagation \cite{he2025dreamstory,NEURIPS2022_960a172b}. 
 % Indeed,  In our case study on the FlintStoneStory sample (shown in Figure~\ref{fig:tteaser}), we observe pronounced cross-frame drifts in terms of human figures and art styles from Frame 1 to Frame 13 across all baselines.
Consequently, existing methods continue to exhibit cross-frame drift, especially in maintaining consistent human figures and visual styles over extended narratives.
This limitation largely stems from the absence of explicit state propagation across frames, leaving identity and stylistic attributes to be repeatedly re-sampled.
% As illustrated in Figure~\ref{fig:introteaser}, such drift becomes increasingly pronounced from Frame~1 to Frame~13 across all baselines on the \textit{FlintStoneStory} sample.
% As illustrated in Figure~\ref{fig:introteaser}, visual drift accumulates as the story unfolds under traditional prompting pipelines, whereas S2ED maintains stable character identity and appearance across frames on the \textit{FlintStoneStory} example.
%As illustrated in Figure~\ref{fig:introteaser}, such drift compounds over long sequences under traditional prompting pipelines, whereas S2ED introduces an explicit, frame-level description interface that preserves character identity and appearance across panels.

In this paper,  we frame story illustration as a \emph{compiler problem} by providing an explicit state interface that {\em decouples narrative reasoning from image synthesis.} Concretely, we propose \emph{Story-to-Executable Descriptions (S2ED)} pipeline that compiles a full story into per-frame \emph{executable descriptions} that downstream text-to-image (T2I) models can follow.
S2ED makes the visual state explicit: \emph{who} appears, \emph{how} they look, \emph{where} they are, and \emph{how} they feel, and carries this state across frames to reduce drift while enabling local, human-editable control without retraining the renderer.
Concretely, this training-free pipeline performs (i) narrative segmentation, (ii) character consistency grounding from a canonical library, and (iii) visual enrichment for layout and affect.

To support systematic evaluation of long-range narrative consistency, we release \emph{Flintstones} (166 stories) and show that S2ED improves cross-frame character alignment and consistency over strong prompting and large-model baselines in both automatic metrics and a controlled human preference study.
We further deploy S2ED in an end-to-end story-to-storybook system featuring fixed IP characters from the Arabic-language Shakoo Maku series (video and details in Supplementary), illustrating its practical potential.

\smallskip

\noindent\textbf{Contributions.}
\begin{itemize}[leftmargin=*]
    \item We introduce S2ED, which compiles stories into executable, frame-level descriptions that explicitly encode and propagate identity, layout, and affect to mitigate cross-frame drift.
    \item  We present a modular workflow for segmentation, grounding, and enrichment that is model-agnostic and supports human-in-the-loop edits without retraining.
    \item We release \emph{Flintstones} and validate S2ED with automatic metrics and a human study, plus an end-to-end deployment demo in Supplementary.
\end{itemize}

% Our contributions are summarized below: 
% \begin{enumerate}[leftmargin=*]
%     \item Define \emph{Story-to-Executable Descriptions (S2ED)} as an intermediate objective bridging narrative text and multi-frame generation.
%     \item Propose a training-free, model-agnostic multi-agent approach for segmentation, grounding, and enrichment.
%     \item Introduce the \emph{FlintStoneStory} dataset for evaluating narrative-to-visual consistency.
%     \item Demonstrate significant gains over strong prompting and direct large-model baselines.
% \end{enumerate}

\begin{figure}[t]
    \centering
    \includegraphics[width=\columnwidth]{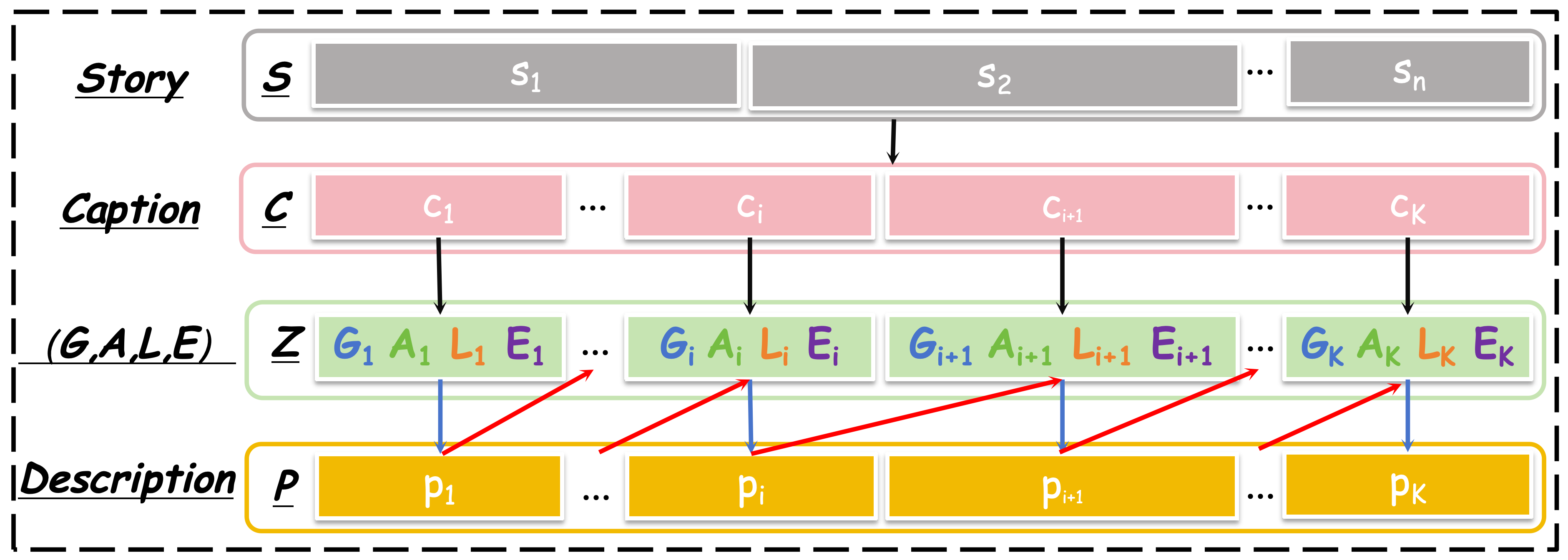}
    \caption{\textbf{S2ED workflow.} 
    Story sentences are segmented into captions, converted into structured 
    states $Z_i$, and combined recursively with prior descriptions to generate 
    Executable Descriptions $p_i$.}
    \label{fig:S2ED-workflow}
\end{figure}

\section{Related Work}

\noindent {\bf Multi-frame Story Illustration:}
Recent story illustration systems based on diffusion models improve narrative coherence but still struggle with cross-frame visual consistency. StoryDiffusion~\cite{NEURIPS2024_c7138635} and Characterfactory~\cite{wang2025characterfactory} enhance consistency via feature sharing or regularization, but typically introduce additional training or rely on constrained setups. In contrast, training-free pipelines such as Infinite-Story~\cite{park2025infinite} and One-Prompt-One-Story~\cite{liu2025one} show that structured prompting can reduce drift without retraining; however, they largely operate with global prompts or handcrafted mechanisms and provide limited \emph{frame-level} control over identity, layout, and affect.

\smallskip

\noindent {\bf Controllability in T2I:}
A complementary line of work studies subject control and personalization in text-to-image generation. Fine-tuning approaches (e.g., DreamBooth~\cite{ruiz2023dreambooth}, LoRA~\cite{hu2021}) can stabilize appearance but require instance-specific data and compute, while token/embedding-based methods (e.g., Textual Inversion~\cite{gal2022image}) encode identity cues yet do not explicitly maintain spatial or affective consistency across a multi-frame sequence. These gaps motivate methods that expose and propagate an explicit per-frame visual state rather than only editing a single global prompt.

\smallskip

\noindent{\bf Agentic Decomposition for Structured Generation:}
Agentic prompting frameworks such as ReAct~\cite{yao2023react} demonstrate that decomposing complex tasks into iterative steps can improve robustness; S2ED adopts this principle to construct and propagate structured visual state for multi-frame story illustration.

\section{S2ED Workflow}
Stories are written in natural language, whereas multi-frame illustration requires
structured, persistent guidance across frames. We formulate S2ED as a
\emph{story-to-structure compilation} pipeline that converts a story into
frame-level \emph{executable descriptions} for an arbitrary renderer
(See Fig.~\ref{fig:S2ED-workflow}).

Let a story $S$ be a sequence of sentences and let $K$ be the desired number of frames.
% S2ED first produces $K$ frame-aligned captions
% $C=(c_1,\ldots,c_K)$, where $c_i$ summarizes the content intended for frame $i$.
% For each caption, S2ED constructs a structured state
% \[
% Z_i=(G_i,A_i,L_i,E_i),
% \]
S2ED first produces $K$ frame-aligned captions
$C=(c_1,\ldots,c_K)$ to capture the intended semantics of each frame.
Rather than being rendered directly, each caption $c_i$ is compiled into a structured visual state:
\[
Z_i=(G_i,A_i,L_i,E_i),
\]

where $G_i$ is a character registry (from $\mathcal{G}$), $A_i$ encodes canonical
appearance attributes (from $\mathcal{A}$), $L_i$ specifies spatial layout, and
$E_i$ captures affective cues. This state exposes the consistency-critical
information needed to preserve identity, composition, and mood across frames.

Crucially, frames are not generated independently. The executable description for
frame $i$ is updated recurrently from the previous description and the new state:
$p_i = f(p_{i-1}, Z_i)$,
so that prior commitments (e.g., identity and style) are carried forward unless
the current caption indicates a change.

A solution is \emph{valid} if it satisfies: \textit{(i) local alignment}, where
$p_i$ matches the semantics of $c_i$; and \textit{(ii) global consistency}, where
the state sequence preserves invariants across frames unless supported by the
narrative (e.g., stable appearance for persistent entities, coherent layout
changes, and consistent affect progression). Finally, each frame is rendered by
any text-to-image model: $\mathrm{Image}_i=\mathrm{T2I}(p_i)$.

\section{Methodology}

\begin{figure*}[t]
    \centering
    \includegraphics[width=0.75\textwidth]{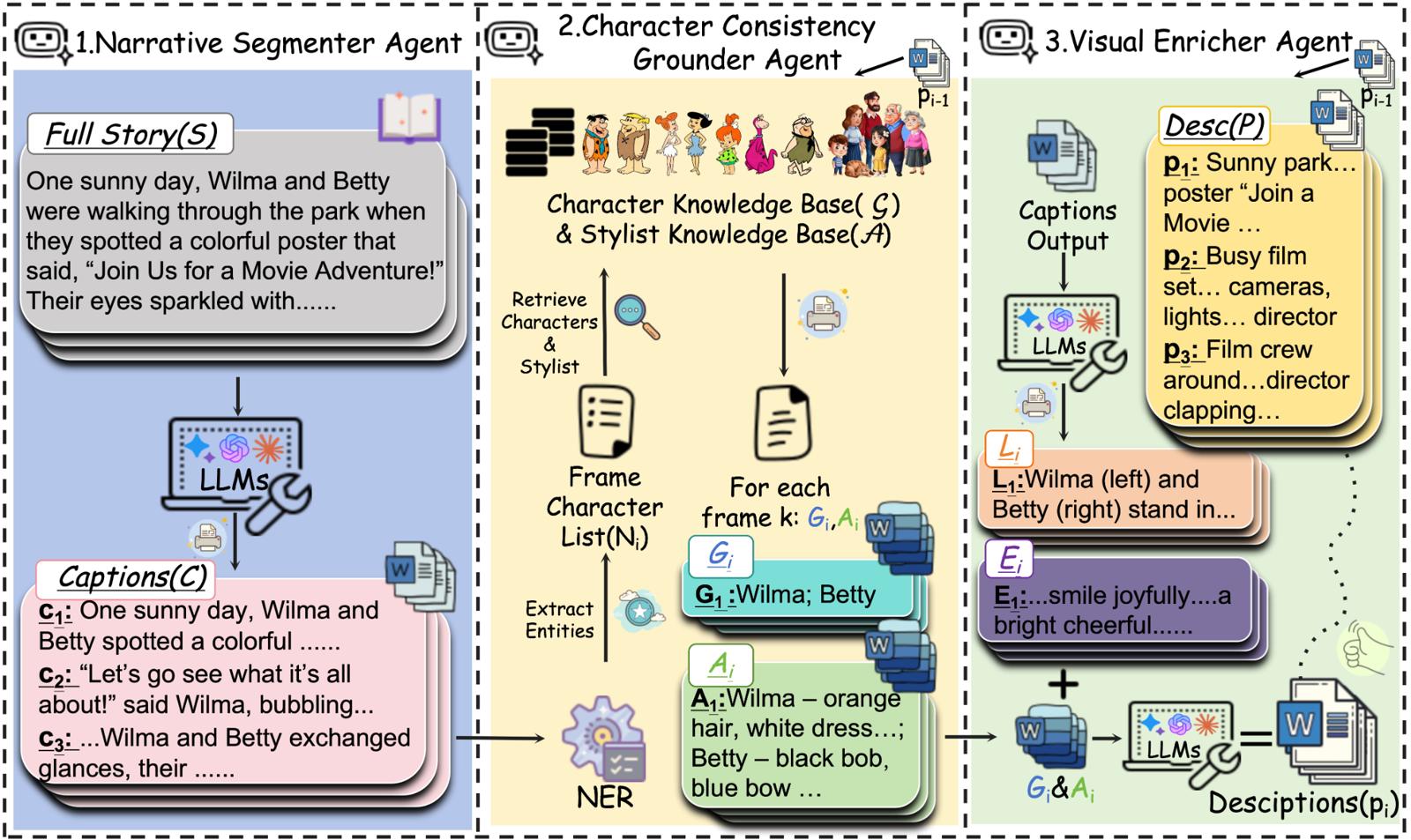}
    \caption{\textbf{Overview of S2ED.}
    The \emph{Narrative Segmenter} produces frame-level captions, the 
    \emph{Consistency Grounder} extracts character and appearance attributes 
    from the caption and global knowledge bases, and the \emph{Visual Enricher} 
    integrates layout and affect cues to produce Executable Descriptions for 
    T2I rendering.}
    \label{fig:S2ED-mechanism}\vspace*{-0.1cm}
\end{figure*}

%S2ED is implemented as a lightweight multi-agent system operating at the prompt layer.
%Given the formal workflow described in Section~III, the system instantiates this process through three specialized components that coordinate narrative
%segmentation, consistency grounding, and visual enrichment.
\noindent {\bf A. Overview:} S2ED is implemented as three prompt-layer agents (Fig.~\ref{fig:S2ED-mechanism}):
a \emph{narrative segmenter} that produces frame-aligned captions, a
\emph{consistency grounder} that instantiates character identity/appearance, and a
\emph{visual enricher} that adds layout and affect and composes the final executable
description.
%Figure~\ref{fig:S2ED-mechanism} illustrates the interaction among these components and their roles in constructing Executable Descriptions.

%Given a story $S$ and target length $K$, the Segmenter produces captions $C=(c_1,\ldots,c_K)$, where each $c_i$ summarizes the content intended for frame $i$. We use deterministic LLM decoding for reproducibility (details in Supplementary).

\smallskip

\noindent {\bf B. Narrative Segmenter:} The \emph{narrative segmenter} partitions the story $S$ into $K$ frame-aligned visual units, producing a sequence of captions $(c_1,\ldots,c_K)$.
where each caption $c_i$ summarizes the characters, actions, and scene 
elements relevant to frame $i$. An LLM is prompted with few-shot instructions 
to maintain discourse order, avoid introducing entities not present in the 
story, respect event boundaries, and compress multi-sentence observations when 
appropriate. Decoding is deterministic (temperature 0, top-$p{=}0$). 
%
% Each caption $c_i$ serves as the sole textual input for extracting the 
% structured state $Z_i$, while also informing which commitments in 
% $p_{i-1}$ must be preserved or updated.
Each caption $c_i$ serves as the sole \emph{new} textual input for constructing the structured state $Z_i$, while the previous description $p_{i-1}$ provides inherited commitments that are selectively preserved or updated.

\smallskip

\noindent {\bf C. Character Consistency Grounder:} 
The \emph{consistency grounder} constructs the character-related components
$(G_i, A_i)$ of the state $Z_i$ using the current caption $c_i$, global
identity/stylist knowledge bases, and \emph{prompt-carried state} $p_{i-1}$ from the
previous frame.

First, we extract and canonicalize the set of characters mentioned in $c_i$:
$N_i = \mathrm{resolve}(\textsc{ExtractEntities}(c_i))$. 
For each character $e \in N_i$, the grounder retrieves the canonical identity
record from the global character knowledge base:
\[
G_i = \{\, (e, G(e)) \mid e \in N_i \,\}.
\]
Next, to establish per-frame style, the Grounder selects attributes from the
global stylist knowledge base $A(e)$, optionally guided by textual cues in
$c_i$ (e.g., outfit changes). We do \emph{not} maintain a separate learned
latent memory or an external cross-frame identity tracker/table. Instead, S2ED
propagates state explicitly at the prompt layer: when $p_{i-1}$ contains a
visual commitment for character $e$ (e.g., hair style, clothing, distinctive
traits), the Grounder preserves that commitment by conditioning style selection
on the content of $p_{i-1}$. In this way, identity is carried forward as
explicit, editable textual constraints rather than as hidden model state.
Formally, we implement this prompt-carried propagation as:
\[
A_i(e) = \textsc{SelectStyle}(A(e),\, c_i,\, p_{i-1}),
\]
where \textsc{SelectStyle} prioritizes attributes already specified in
$p_{i-1}$ unless $c_i$ provides evidence of an intentional change (e.g., an
explicit outfit change).
The grounder outputs the character registry and stylist table as part of $Z_i$.

\smallskip

\noindent {\bf D. Visual Enricher:}
The \emph{Visual Enricher} constructs the layout and affect fields $(L_i,E_i)$ 
and composes the final Executable Description. Unlike recurrent models that 
store explicit spatial or emotional states, S2ED derives all previously 
established commitments from the textual content of $p_{i-1}$.

New layout and affect cues are extracted from $c_i$:
{\small \[
L_i^{\text{new}} = \textsc{LayoutCues}(c_i), 
\qquad
E_i^{\text{new}} = \textsc{AffectCues}(c_i).
\]}

These cues are integrated with any implicit commitments contained in the 
previous description:
{\small \[
L_i = \textsc{Integrate}(L_i^{\text{new}},\, p_{i-1}), 
\text{\ \ }
E_i = \textsc{Integrate}(E_i^{\text{new}},\, p_{i-1}).
\]}

% The full structured state for frame $i$ is thus:
% \[
% Z_i = (G_i, A_i, L_i, E_i).
% \]

% Finally, the Executable Description is generated by combining the inherited 
% constraints from $p_{i-1}$ with the freshly constructed state:
% \[
% p_i = f(p_{i-1}, Z_i),
% \]
% where $p_{i-1}$ provides global consistency and $Z_i$ injects frame-specific 
% narrative structure. The description $p_i$ is then fed into any pretrained 
% text-to-image model to produce $\mathrm{Image}_i$.
Finally, we compose the executable description by combining inherited constraints
from $p_{i-1}$ with the structured state $Z_i$, yielding $p_i$ for rendering.

%The full structured state for frame $i$ thus aggregates character identity,
%ppearance attributes, spatial layout, and affective cues.
%Finally, the executable description is generated by combining the inherited
%constraints from $p_{i-1}$ with the newly constructed state, yielding a
%frame-specific prompt that preserves global consistency while incorporating
%local narrative information. The resulting description $p_i$ is then fed into
%any pretrained text-to-image model to produce $\mathrm{Image}_i$.

The prompt specifications, input/output formats, and execution constraints
for the three modules above are documented
in the Supplementary Material (Section~3).

\begin{figure*}[t]
  \centering

  % 第一张图
  \includegraphics[width=0.67\textwidth]{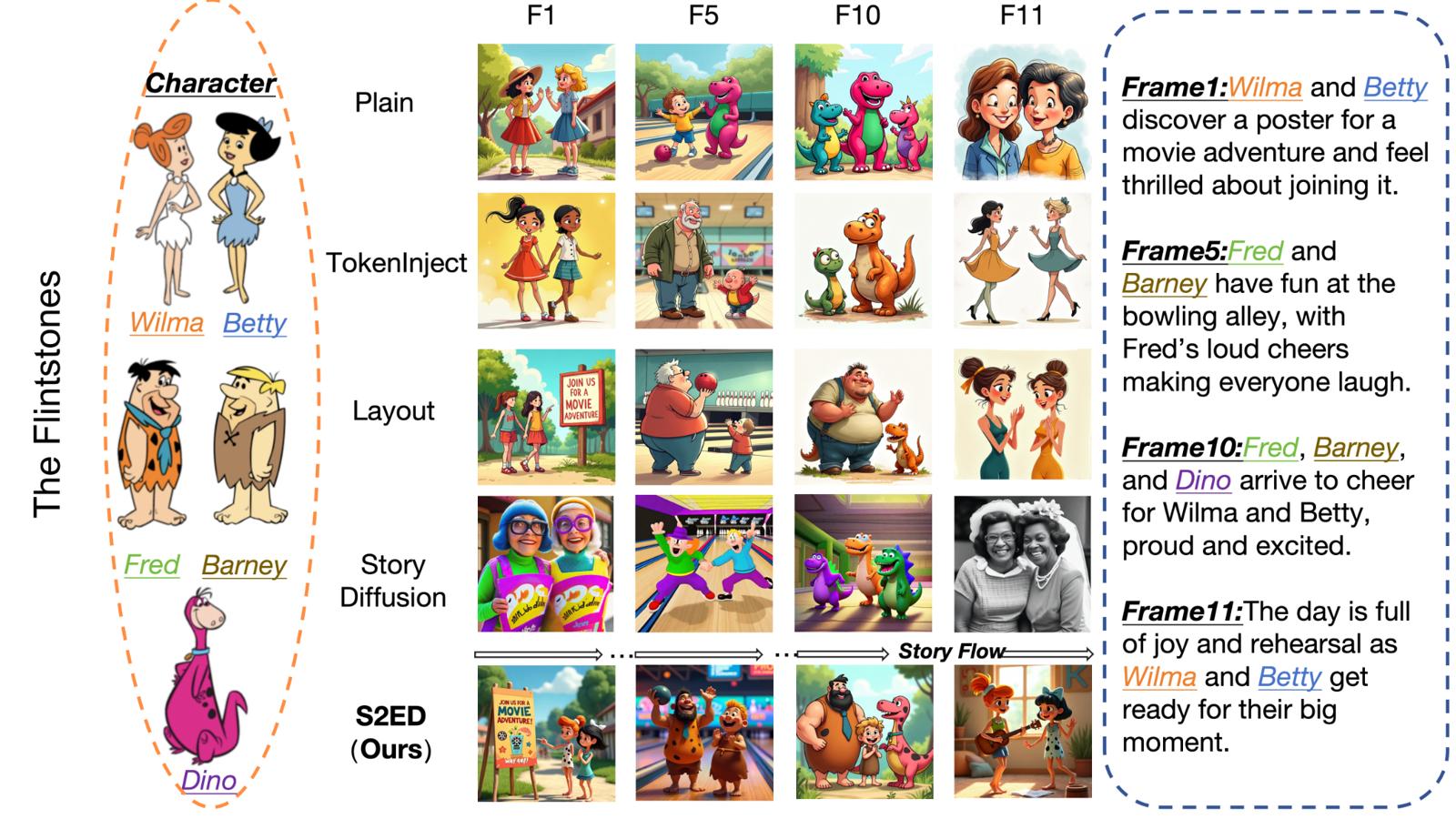}
  \vspace{-1mm}

  % 第二张图
  \includegraphics[width=0.67\textwidth]{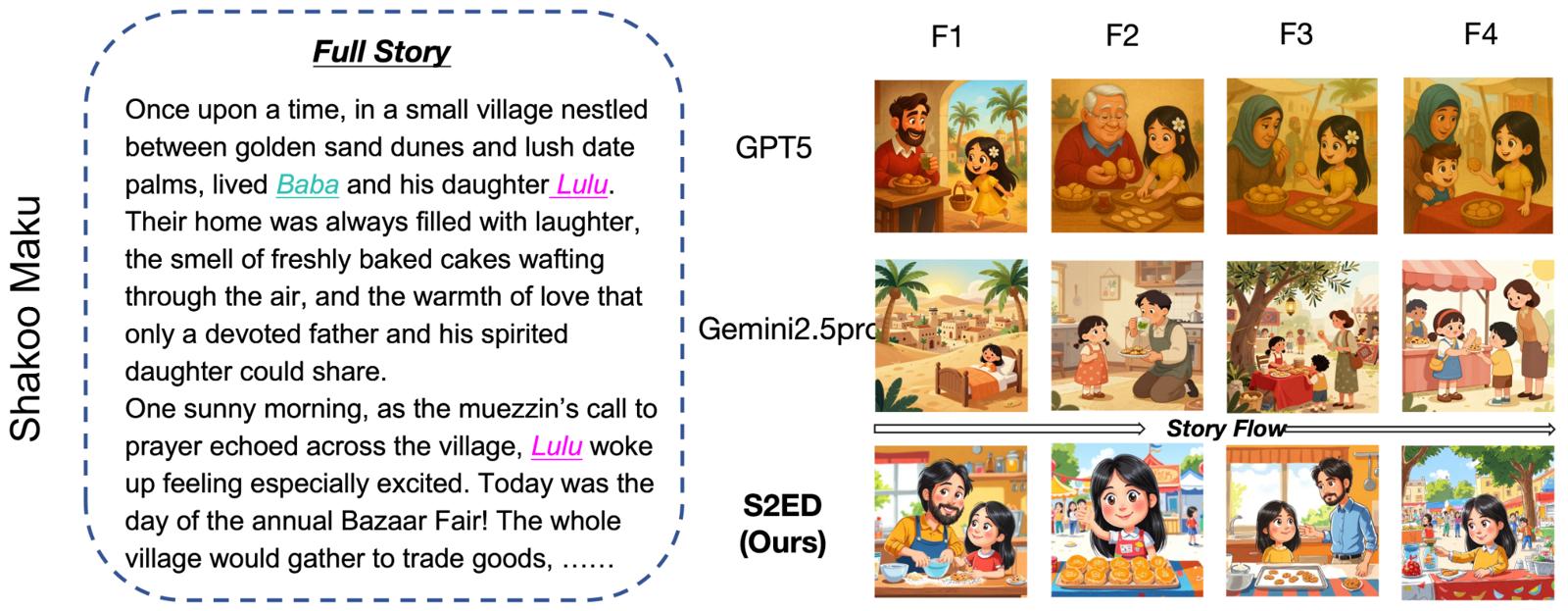}

    \caption{\textbf{Qualitative overview of S2ED.}
Top: Results on the Flintstones dataset comparing S2ED with prompting baselines (Plain, TokenInject, Layout) and StoryDiffusion across multiple frames (1, 5, 10, 11).
Bottom: Results on the Shakoo Maku dataset showing end-to-end story-to-image generation (Frames 1--4), comparing GPT-5, Gemini-2.5 Pro, and S2ED.}

  \label{fig:tteaser}
  \vspace{-3mm}
\end{figure*}

\section{Experiments}

%We evaluate S2ED on story illustration using automatic metrics, story-level coherence measures, human evaluations, and cross-domain generalization. The study investigates whether multi-agent prompt decomposition enhances visual and narrative consistency over strong baselines, how GALE-based executable descriptions improve identity preservation and spatial stability, the degree to which automatic metrics align with human perception, and the ability of S2ED to generalize beyond a single IP or visual style.

We evaluate whether S2ED improves (i) cross-frame identity and layout consistency and (ii) story-to-frame event alignment over strong prompting baselines, and whether the gains transfer across IP/styles. We report automatic metrics and human judgments under a fixed T2I backend.

\subsection{Dataset}

\paragraph{Flintstones}
We release the \textit{Flintstones} dataset
based on the classic animated series \textit{The Flintstones}, which features
recurring characters and visually distinctive scenes.
We collect episode synopses from public sources (e.g., Wikipedia) and rewrite
them into concise, child-friendly narratives, with each sentence describing a
clear and visualizable event. The rewriting is conducted by the authors and manually verified to ensure
narrative coherence and suitability for visual illustration.
% Each story is paired with a structured character library encoding canonical
% attributes such as hairstyle, clothing, and accessories.
The dataset includes a global structured character library that encodes canonical appearance attributes such as hairstyle, clothing, and accessories.
The dataset contains 166 stories (25-38 sentences; 3-5 main characters), producing $\sim$6 frames/story after segmentation, and includes a character library with canonical appearance attributes. \emph{Availability \& license:} CC~BY~4.0 at
\href{https://anonymous.4open.science/r/FlintStoneStory-Dataset-S2ED/}{anonymous repository}. An example story and corresponding workflow is included in the Supplementary Material (Section~1).

%\noindent {\em Availability \& License:}
%The \textit{FlintStoneStory} dataset is released under the Creative Commons
%Attribution 4.0 International License (CC~BY~4.0).
%It is publicly available for research and benchmarking purposes at an anonymized repository: \href{https://anonymous.4open.science/r/FlintStoneStory-Dataset-S2ED/}{https://anonymous.4open.science/r/FlintStoneStory-Dataset-S2ED/}.

\paragraph{Shakoo Maku} 
We additionally evaluate S2ED on the \textit{Shakoo Maku} dataset, a
collection of short illustrated stories featuring recurring characters
such as \textit{Lulu}, \textit{Zain}, \textit{Baba}, and \textit{Mama}.
This dataset draws inspiration from narratives and character
constructions found in the Shakoo Maku edutainment universe (\url{https://www.shakoomaku.com/}).% The original Shakoo Maku brand and story world can be accessed at: \url{https://www.shakoomaku.com/}.
The stories in Shakoo Maku cover everyday scenarios and family-oriented
narratives, providing a stylistically different but structurally stable
environment compared to Flintstones. %Because the characters appear consistently across many stories, Shakoo Maku offers a complementary test bed for evaluating identity preservation and cross-story coherence.
%
%We apply the same pipeline as before: segmenting each story into captions, generating executable descriptions with S2ED, and rendering images using a pretrained diffusion model.
%
A full example story and corresponding workflow is provided in the Supplementary Material (Section~2).

\subsection{Baselines}
All methods use FLUX-1 (dev, FP8) as the T2I backend.
 \textbf{Full-story baselines} generate per-frame prompts
from the full story using (1) GPT-5 and (2) Gemini-2.5 Pro. \textbf{Caption-based baselines}
assume frame captions: (3) \textsc{PlainPrompt} (raw captions), (4) \textsc{TokenInject}
(caption + appearance), and (5) \textsc{LayoutPrompt} (caption + layout/affect).
We exclude per-identity fine-tuning (e.g., DreamBooth) as it targets single-subject customization
and requires additional optimization per character. For caption-based baselines, we use the same captions produced by our segmenter to control quality.

%\subsection{Baselines}
%We compare the S2ED framework against two groups of baselines, all using the same T2I model backend:

%\paragraph{Full-story baselines}  
%These methods use the full story as input and rely on LLMs to generate per-frame prompts:

%\begin{itemize}[leftmargin=*]
%    \item \textbf{GPT5}: GPT-5 to transform the full story into per-frame prompts. This provides global reasoning but lacks explicit mechanisms for identity or layout consistency.
%    \item \textbf{Gemini-2.5 Pro}: Uses Gemini-2.5 Pro to generate per-frame prompts from the full story. Similar to GPT5, it leverages global context but has no structured consistency control.
%\end{itemize}

%\paragraph{Caption-based baselines.}  
%These methods assume that frame-level captions are available and operate on them.

%\begin{itemize}[leftmargin=*]
%    \item \textbf{PlainPrompt}: Directly uses raw captions as prompts, with no segmentation, grounding, or enrichment.
%    \item \textbf{TokenInject}: Adds character appearance attributes to captions, but omits layout or emotion.
%    \item \textbf{LayoutPrompt}: Adds layout, background, and emotion cues, but does not enforce character identity.
%\end{itemize}
\noindent {\em Paradigm-level comparison:}
Unlike S2ED, which takes a full story as input and compiles it into frame-level executable descriptions, most SOTA systems (e.g., StoryDiffusion \cite{NEURIPS2024_c7138635}) assume per-frame captions are already provided and use dedicated diffusion backbones, making them not directly comparable to our fixed-backend prompting baselines. We therefore report StoryDiffusion only as a reference by feeding it the same captions (and character library when supported) to contrast training-free prompt-carried state propagation with training-based consistency modeling.

%We do not include other fine-tuned personalization methods such as DreamBooth, as they require per-identity optimization and are designed for single-subject customization rather than multi-character story illustration. Such methods do not explicitly address cross-frame identity propagation or long-range narrative coherence, which are central to the task considered here.

%We additionally report results from StoryDiffusion as a reference
%training-based consistency method. Due to its architectural design,
%StoryDiffusion employs a dedicated diffusion backbone and therefore does
%not share the same T2I backend as the prompting-based baselines above.
%Accordingly, it is used for paradigm-level comparison rather than as a strictly controlled backbone-fixed baseline. To ensure semantic comparability, we provide StoryDiffusion with the same frame-level captions and the same canonical character library whenever applicable. This comparison is intended to contrast prompt-layer, training-free state propagation with training-based consistency modeling, rather than to compare raw image quality across backbones.

\smallskip

\subsection{Evaluation Metrics}
We assess story-level consistency with automatic metrics and a human study.

\begin{itemize}[leftmargin=*]
    \item \textbf{CLIPScore.}
    Semantic alignment via cosine similarity between CLIP image/text embeddings~\cite{hessel2021clipscore}.

    % \item \textbf{Character F1 (Char-F1).}
    % Character identity consistency using the recognition protocol of~\cite{rahman2023make} (cf.\ Eq.~(5) in the survey).

    \item \textbf{Character F1 (Char-F1).}
    Character identity consistency using the recognition protocol of~\cite{rahman2023make} .

    \item \textbf{Frame Accuracy (F-Acc).}
    Event alignment following~\cite{maharana2022storydall}, computed by matching predicted frame labels to action labels extracted from captions.

    \item \textbf{Spatial Consistency (mAP).}
    Layout/spatial grounding measured by COCO-style mAP under IoU thresholds~\cite{lin2014microsoft,li2022grounded,li2023gligen}.

    \item \textbf{Human Evaluation.}
    We conduct a human study using 5-point Mean Opinion Scores and pairwise preference tests, following practice for temporal and  stylistic coherence.
\end{itemize}

\subsection{Results on Flintstones}

We report results on the Flintstones dataset.
Table~\ref{tab:flintstones-main} shows that S2ED outperforms all baselines on all
four automatic metrics. S2ED achieves the highest CLIPScore, indicating better semantic alignment between the generated frames and the underlying story text.
The gains in Char-F1 are particularly notable: compared to the strongest
baseline, S2ED improves character identity consistency by more than
7 F1 points, reflecting fewer appearance changes and better preservation
of clothing, props, and other visual attributes. 
On the temporal side, S2ED yields the best Frame Accuracy (F-Acc),
indicating that generated action sequences follow the intended
plot faithfully. Finally, S2ED achieves the highest mAP, suggesting
that spatial layouts and object placements remain stable across
frames than competing methods.

%We first report results on the FlintStoneStory dataset, which is specifically constructed to test identity preservation, event alignment, and spatial stability under a fixed IP. The quantitative comparison in Table~\ref{tab:flintstones-main} shows that S2ED consistently outperforms all baselines across all four automatic metrics.

\begin{table}[h]
\centering
\caption{Quantitative results on the Flintstones dataset.
Higher is better for all metrics.}
\label{tab:flintstones-main}
\resizebox{\linewidth}{!}{
\begin{tabular}{lcccc}
\toprule
\textbf{Method} &
\textbf{CLIPScore} $\uparrow$ &
\textbf{Char-F1} $\uparrow$ &
\textbf{F-Acc} $\uparrow$ &
\textbf{mAP} $\uparrow$ \\
\midrule
PlainPrompt     & 0.265 & 0.58 & 0.41 & 0.32 \\
TokenInject     & 0.272 & 0.62 & 0.43 & 0.34 \\
LayoutPrompt    & 0.279 & 0.60 & 0.47 & 0.42 \\
GPT5            & 0.284 & 0.63 & 0.48 & 0.39 \\
Gemini-2.5 Pro  & 0.286 & 0.64 & 0.49 & 0.40 \\
\textbf{S2ED (Ours)} &
\textbf{0.298} & \textbf{0.71} & \textbf{0.56} & \textbf{0.51} \\
\bottomrule
\end{tabular}
}
\end{table}

% On the temporal side, S2ED also yields the best Frame Accuracy (F-Acc),
% showing that the sequence of generated actions follows the intended
% plot more faithfully. Finally, S2ED reaches the highest mAP, suggesting
% that spatial layouts and object placements remain more stable across
% frames than in competing methods.

Figure~\ref{fig:tteaser} shows qualitative examples. PlainPrompt changes character outfits and
occasionally produces actions that do not match the story. TokenInject
keeps some appearance details but fails to maintain structure in cluttered scenes. LayoutPrompt yields better spatial organization but
often loses identity. GPT5 and Gemini-2.5 Pro benefit from full-story context but still show frame-to-frame drift. In contrast, S2ED produces
sequences where identity, actions, and layouts evolve coherently over the entire story.

We include StoryDiffusion as a representative consistency method that relies on model training for qualitative comparison.
While StoryDiffusion benefits from an identity aware diffusion backbone, it can exhibit appearance drift or over regularization across longer sequences.
In contrast, S2ED achieves comparable or stronger identity stability using a prompt level mechanism without additional model training.

% \begin{figure}[h]
% \centering
% \includegraphics[width=0.47\textwidth]{figures/flintstones_qualitative.png}
% \caption{Qualitative examples on FlintStoneStory. Compared to baselines,
% S2ED maintains more stable identities, actions, and spatial layouts
% across frames.}
% \label{fig:flintstones-qual}
% \end{figure}

\subsection{Results on Shakoo Maku}

%We further evaluate S2ED on the Shakoo Maku dataset to test whether the framework generalizes to a different narrative domain, additional story themes, and a separate set of recurring characters. %(e.g., Lulu, Zain, Baba, Mama). 
Table~\ref{tab:shakoomaku-main} summarizes the results on the Shakoo Maku dataset.
Although the visual style and story structure differ noticeably from
Flintstones, S2ED again achieves the best performance across
all automatic metrics. 
Compared to Flintstones, all baselines show a larger performance
drop due to higher stylistic diversity and more open-ended story
structure. In contrast, S2ED maintains a stable advantage, improving
character consistency (+7 Char-F1 points), event alignment (+6 F-Acc
points), and spatial grounding (+6 mAP points) over the strongest
baseline. These gains demonstrate that S2ED-based executable descriptions
provide robust cross-domain generalization, even when the stories differ
from those used to construct the character libraries.

\begin{table}[h]
\centering
\caption{Quantitative results on the Shakoo Maku dataset.
Higher is better for all metrics.}
\label{tab:shakoomaku-main}
\resizebox{\linewidth}{!}{
\begin{tabular}{lcccc}
\toprule
\textbf{Method} &
\textbf{CLIPScore} $\uparrow$ &
\textbf{Char-F1} $\uparrow$ &
\textbf{F-Acc} $\uparrow$ &
\textbf{mAP} $\uparrow$ \\
\midrule
PlainPrompt     & 0.241 & 0.54 & 0.38 & 0.29 \\
TokenInject     & 0.250 & 0.58 & 0.40 & 0.31 \\
LayoutPrompt    & 0.258 & 0.56 & 0.44 & 0.38 \\
GPT5            & 0.264 & 0.59 & 0.45 & 0.34 \\
Gemini-2.5 Pro  & 0.267 & 0.60 & 0.46 & 0.36 \\
\textbf{S2ED (Ours)} &
\textbf{0.283} & \textbf{0.67} & \textbf{0.52} & \textbf{0.44} \\
\bottomrule
\end{tabular}
}
\end{table}

On the qualitative example in Fig.~\ref{fig:tteaser}, baselines exhibit similar identity and continuity issues, while
S2ED remains consistently stable and story-faithful, demonstrating cross-domain generalization.

A similar trend is observed when comparing against StoryDiffusion, suggesting that S2ED’s explicit state propagation generalizes beyond training-based consistency mechanisms.

%Figure~\ref{fig:tteaser} illustrates qualitative examples.
%In the example in Fig.~\ref{fig:tteaser},  PlainPrompt and TokenInject lose character tracking when
%style shifts occur across frames, while LayoutPrompt improves structure
%but struggles with identity drift. GPT5 and Gemini-2.5 Pro produce more
%coherent narratives but still exhibit inconsistencies in character
%appearance and action continuity. S2ED produces the most stable and
%story-faithful sequences across the full Shakoo Maku dataset.

% \begin{figure}[h]
% \centering
% \includegraphics[width=0.47\textwidth]{figures/shakoomaku_qualitative.png}
% \caption{Qualitative examples on Shakoomaku. S2ED generalizes across
% different characters and styles, maintaining identity and narrative
% coherence more effectively than baselines.}
% \label{fig:shakoomaku-qual}
% \end{figure}

\subsection{Human Evaluation}

To complement automatic metrics, we conduct a human study with {\em 20 annotators} with
prior experience in visual storytelling or image assessment. Each annotator
rates 10 randomly sampled stories generated by all methods on three criteria:
\textit{character consistency}, \textit{story relevance}, and \textit{overall visual quality},
using a 5-point Likert scale (1 = poor, 5 = excellent). Inter-rater agreement is
high (Fleiss's $\kappa = 0.72$) and internal reliability is strong (Cronbach's
$\alpha > 0.8$).

%To complement the automatic metrics, we conduct a human evaluation with
%20 annotators who have prior experience with visual storytelling or
%image assessment. Each annotator is shown a randomly sampled set of
%stories generated by all methods and is asked to rate three aspects:
%\textit{character consistency}, \textit{story relevance}, and
%\textit{overall visual quality}. Ratings follow a standard 5-point
%Likert scale (1 = poor, 5 = excellent). Inter-rater agreement is high
%(Fleiss' $\kappa = 0.72$) and internal reliability is strong
%(Cronbach's $\alpha > 0.8$).

Table~\ref{tab:human-study} reports mean opinion scores (MOS). S2ED achieves the
highest MOS on all criteria, with the largest margin on character consistency,
indicating substantially fewer identity and appearance mismatches.

%Table~\ref{tab:human-study} summarizes the results. S2ED receives the
%highest mean opinion scores across all dimensions, demonstrating
%clear human preference for S2ED-based executable descriptions. In
%particular, S2ED improves character consistency by a wide margin,
%reflecting significantly fewer identity or appearance mismatches.

\begin{table}[h]
\centering
\caption{Human evaluation results (mean $\pm$ std) on a 5-point Likert scale.
Higher is better.}
\label{tab:human-study}
\resizebox{\linewidth}{!}{
\begin{tabular}{lccc}
\toprule
\textbf{Method} &
\textbf{Char. Consistency} $\uparrow$ &
\textbf{Story Relevance} $\uparrow$ &
\textbf{Visual Quality} $\uparrow$ \\
\midrule
PlainPrompt     & 2.3 $\pm$ 0.7 & 2.5 $\pm$ 0.6 & 2.7 $\pm$ 0.7 \\
TokenInject     & 2.8 $\pm$ 0.6 & 2.9 $\pm$ 0.6 & 3.0 $\pm$ 0.6 \\
LayoutPrompt    & 3.1 $\pm$ 0.6 & 3.0 $\pm$ 0.6 & 3.3 $\pm$ 0.7 \\
GPT5            & 3.2 $\pm$ 0.5 & 3.3 $\pm$ 0.6 & 3.4 $\pm$ 0.5 \\
Gemini-2.5 Pro  & 3.3 $\pm$ 0.5 & 3.4 $\pm$ 0.6 & 3.5 $\pm$ 0.5 \\
\textbf{S2ED}   & \textbf{4.4 $\pm$ 0.4} &
\textbf{4.5 $\pm$ 0.4} &
\textbf{4.3 $\pm$ 0.5} \\
\bottomrule
\end{tabular}
}
\end{table}

Beyond MOS, we run pairwise preference tests in which annotators choose the
better sequence for a target criterion. S2ED is preferred in 82\% of comparisons
for character consistency, 78\% for story relevance, and 74\% for visual quality,
confirming that automatic improvements translate into perceptual gains.

The questionnaire, instructions, and rating scales are in Supplementary
Material (Sec.~4). We also report Spearman correlations between automatic
metrics and human scores in Supplementary (Sec.~5).

\subsection{Ablation Study}
We ablate S2ED on Flintstones to quantify the contribution of each 
component in $Z_i=(G_i,A_i,L_i,E_i)$. Starting from the full model, we remove one component at a time
and keep all other settings fixed:

%We conduct an ablation study on FlintStoneStory to quantify the contribution of each component in the  state
%$Z_k = (G_k, A_k, L_k, E_k)$, where $G_k$ denotes grounded character
%identities, $A_k$ encodes appearance attributes, $L_k$ represents
%spatial layout, and $E_k$ captures emotional and affective cues.
%Starting from the full S2ED model, we remove one component at a time and keep all other settings unchanged.

\begin{itemize}[leftmargin=*]
    \item \textbf{w/o $G$ (identity grounding):} character IDs are not
    explicitly tracked across frames; prompts only contain local
    mentions without cross-frame bindings.
    \item \textbf{w/o $A$ (appearance attributes):} clothing, hairstyle,
    and prop descriptors are omitted, while identity names and roles are
    preserved.
    \item \textbf{w/o $L$ (layout cues):} spatial relationships,
    relative positions, and scene topology are removed from the
    executable descriptions.
    \item \textbf{w/o $E$ (emotion cues):} affective states and
    expression-related descriptors are excluded from the prompts.
\end{itemize}

Table~\ref{tab:ablation} reports the results. Removing any component degrades
performance, with distinct effects across metrics. Dropping $G$ causes the
largest Char-F1 decrease, underscoring its role in identity preservation.
Removing $A$ further reduces Char-F1 and CLIPScore, suggesting that detailed
appearance cues support both identity stability and text--image alignment.
Omitting $L$ primarily hurts mAP and F-Acc, confirming that explicit spatial
structure improves scene topology and event realization. Finally, removing $E$
yields a moderate decline across metrics, indicating that affective cues help
refine narrative and visual coherence. Overall, all components provide
complementary benefits to story-level consistency.

%Table~\ref{tab:ablation} reports the results. Each GALE component contributes to a different aspect of consistency, and removing any of them degrades performance.

\begin{table}[h]
\centering
\caption{Ablation study on the Flintstones dataset.
Higher is better for all metrics.}
\label{tab:ablation}
\resizebox{\linewidth}{!}{%
\begin{tabular}{lcccc}
\toprule
\textbf{Variant} &
\textbf{CLIPScore} $\uparrow$ &
\textbf{Char-F1} $\uparrow$ &
\textbf{F-Acc} $\uparrow$ &
\textbf{mAP} $\uparrow$ \\
\midrule
S2ED (Full)     & {\bf 0.298} & {\bf 0.71} & {\bf 0.56} & {\bf 0.51} \\
w/o $G$ (identity)  & 0.286 & 0.61 & 0.53 & 0.47 \\
w/o $A$ (appearance) & 0.292 & 0.66 & 0.54 & 0.49 \\
w/o $L$ (layout)    & 0.290 & 0.68 & 0.51 & 0.43 \\
w/o $E$ (emotion)   & 0.294 & 0.69 & 0.52 & 0.48 \\
\bottomrule
\end{tabular}
}
\end{table}

\section{Limitation}
\begin{figure}[t]
\centering
\begin{subfigure}{0.35\linewidth}
  \includegraphics[width=\linewidth]{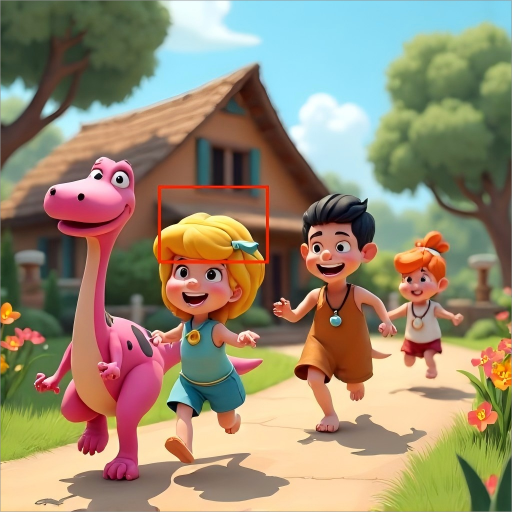}
  \caption{Multi-entity interference with identity drift.}
\end{subfigure}\qquad  
\begin{subfigure}{0.35\linewidth}
  \includegraphics[width=\linewidth]{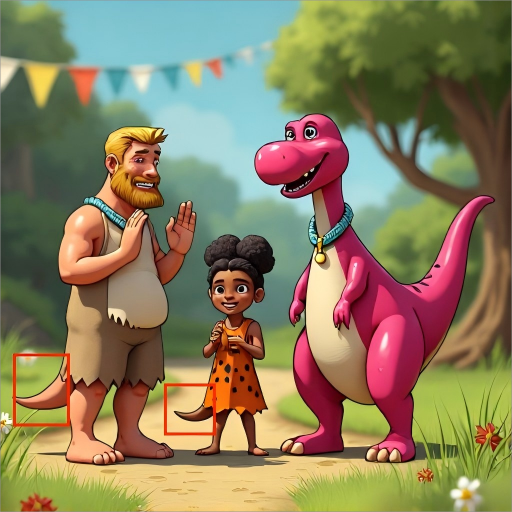}
  \caption{Attribute leakage \& identity confusion}
\end{subfigure}
\caption{Representative failure cases in S2ED. 
(a) Multi-entity interference. 
(b) Attribute leakage and identity confusion.}
\label{fig:failure}
\end{figure}

% We introduced S2ED, a training-free, prompt-layer workflow that converts stories into executable descriptions. 
% Using a modular pipeline, S2ED yields controllable prompts that improve cross-frame consistency and story alignment in multi-frame illustration while remaining compatible with off-the-shelf T2I renderers.  
Several {\em limitations} remain:

\noindent\textbf{(a) Multi-entity interference:}
Crowded frames split attention and weaken identity cues, causing proportion drift and appearance swaps. 
Layout-aware grounding and cross-frame entity tracking could improve robustness  (Fig.~\ref{fig:failure}(a)).
\noindent\textbf{(b) Attribute leakage and identity confusion:}
Overlapping cues can trigger attribute transfer across entities (e.g., non-human parts leaking to humans) during enrichment. 
State locking and attribute-type constraints are needed to prevent non-transferable traits from propagating (Fig.~\ref{fig:failure}(b)). 
\noindent\textbf{(c) Dataset and metric bias:}
Evaluation centers on \textit{Flintstones}, and CLIP-style metrics capture coarse alignment but not narrative coherence or emotional flow. 
Broader datasets and more human-centered measures would strengthen the conclusions.

\bibliographystyle{IEEEbib}
\bibliography{ICME2026/icme2026ref-short}

% \vspace{12pt}
% \color{red}
% IEEE conference templates contain guidance text for composing and formatting conference papers. Please ensure that all template text is removed from your conference paper prior to submission to the conference. Failure to remove the template text from your paper may result in your paper not being published.

\end{document}